\documentclass[final]{cvpr}

\usepackage[T1]{fontenc}
\usepackage{textcomp}
\usepackage[scaled=.92]{helvet}
\usepackage{times}
\usepackage{epsfig}
\usepackage{graphicx}
\usepackage{amsmath}
\usepackage{amssymb}
\usepackage{balance}  
\usepackage{booktabs} 
\usepackage{ragged2e} 
\usepackage{subfig}
\usepackage{adjustbox}
\usepackage{pstricks}
\usepackage{diagbox}
\usepackage{algorithm} 
\usepackage{algorithmic}   
\usepackage{multirow}
\usepackage{cite}
\usepackage{amssymb}
\usepackage{setspace}
\usepackage{ntheorem}
\usepackage[pagebackref=true,breaklinks=true,colorlinks,bookmarks=false]{hyperref}

\newtheorem*{definition}{Definition}

\usepackage{enumitem}
\setenumerate[1]{itemsep=0pt,partopsep=0pt,parsep=\parskip,topsep=5pt}
\setitemize[1]{itemsep=0pt,partopsep=0pt,parsep=\parskip,topsep=5pt}
\setdescription{itemsep=0pt,partopsep=0pt,parsep=\parskip,topsep=5pt}

\def\cvprPaperID{8325} 
\def\confYear{CVPR 2021}

\begin{document}

\title{MinConvNets: A new class of multiplication-less Neural Networks}

\author{Xuecan Yang, Sumanta Chaudhuri, Laurence Likforman, Lirida Naviner\\
LTCI - Télécom Paris, Institut Polytechnique de Paris\\
Palaiseau, France\\
{\tt\small xuecan.yang@telecom-paris.fr}
}

\maketitle

\begin{abstract}
Convolutional Neural Networks have achieved unprecedented success in image classification, recognition, or detection applications. 
However, their large-scale deployment in embedded devices is still limited by the huge computational requirements, 
i.e., millions of MAC operations per layer. 
In this article,  MinConvNets where the multiplications in the forward propagation are approximated by minimum comparator operations are introduced. 
Hardware implementation of minimum operation is much simpler than multipliers.
Firstly, a methodology to find approximate operations based on statistical correlation is presented.
We show that it is possible to replace multipliers by minimum operations in the forward propagation under certain constraints,
i.e. given similar mean and variances of the feature and the weight vectors. 
A modified training method which guarantees the above constraints is proposed. 
And it is shown that equivalent precision can be achieved during inference with MinConvNets
by using transfer learning from well trained exact CNNs.
\end{abstract}

\section{Introduction}

Nowadays, Convolutional Neural Networks (CNNs) are widely deployed in embedded systems and mobile terminals, such as drones, wearable devices. These applications often have real-time constraints and a very tight power budget.
Because of the limited computing resource of embedded systems, the CNNs are always pre-trained offline and then implemented in embedded systems for the inference stage.
However, CNN is still computationally intensive and resource-consuming in the inference stage due to the convolution calculation requiring a large number of multiply-accumulate (MAC) operations.
For example, AlxeNet~\cite{krizhevsky2012imagenet} requires about 666 million MACs operation to process a $227\times 227$ image.
For more complex tasks, such as object detection, the number of multiplication operations is higher.
Multiplication is always more difficult to implement, or time-consuming in computing~\cite{deschamps2012guide},
therefore, many methods are proposed to speed up or compress the network.

\subsection{Related Works} 
\label{subsec:related}
Quantization is a class of methods that make use of low-precision arithmetic~\cite{krishnamoorthi2018quantizing}.
~\cite{diannao} proposes half-precision floating point, 
~\cite{googletpu} uses 8-bit quantized weights, 
and ~\cite{umuroglu2017finn, bnn} propose the use of binarized weights and feature maps. 
Compared with the original neural network encoded in 32-bit,
the quantized network is smaller, 
and fewer-bit multiplication may cost fewer computing resources for some special platform.
Pruning networks is another widely discussed method to compress and speed up the neural network.
Filter pruning methods that rank filters then remove the less important filters are proposed in ~\cite{molchanov2016pruning,li2016pruning,anwar2017structured}.
In weights pruning methods~\cite{han2015learning,srinivas2017training,tartaglione2018learning}, the less important weights in each filter are dropped off and set to 0, so that sparse networks can be built and can be accelerated by sparse matrix calculator. 


Quantization reduces the consumption of each multiplication while pruning reduces the number of multiplications.
However, but both of them are still necessary to calculate the multiplication.
Different from these methods, we abandon multiplication directly.
The comparison operation which is less time-consuming and easier implemented is proposed in our works to replace multiplication in CNNs.

Some works have also proposed multiplication-less networks.
In~\cite{nakahara2019lightweight}, the weight is quantized to 1 bit, which allows multiplication to be converted into addition/subtraction. 
Furthermore, when the activation is also quantized to 1 bit, as introduced in~\cite{rastegari2016xnor,bnn}, multiplication is converted to XNOR logic gate. 
These methods fix one or more multipliers to $\pm 1$, thereby simplifying the calculation.
Moreover, there are also some networks that break away from multiplication.
~\cite{islam2020extending} uses the hit-and-miss transformation composed of comparison and addition to construct a network.
Our work also uses comparison and addition operations, 
but the network with hit-or-miss transformation has brought a considerable loss of accuracy, while the networks in our work have not lost accuracy.

\subsection{Contribution}
We replace multiplication in CNNs with lighter operations.
The main contributions of this work are:
\begin{itemize}
    \item A set of criteria to measure the degree of approximation between two operations is proposed.
    Based on these criteria, an approximate operation to multiplication is proposed for error-tolerant systems.
    \item On top of the proposed approximate operation, an approximate convolutional layer without multiplication is explored to replace the traditional convolutional layer.
    Then we build and train the CNN by using an approximate convolutional layer, named MinConvNets. 
    The benchmark is applied to MinConvNets to show that lossless accuracy can be achieved in image classification tasks.
    \item This work shows that in addition to multiplication, other lighter operations can still effectively extract image features. This provides a new direction for compressing or accelerating CNNs in future works.
\end{itemize}

\subsection{Methodologies and Organization}
\vspace{-0.15cm}
%
%
Firstly, we discuss the approximate calculation of multiplication in section~\ref{sec:approx}.
In this work, we define similar operations and approximate operations, which can generate results approximate to multiplication under certain constraints, 
such as the constraints to expected value or variance of multiplication operands.
Approximate operations are based on similar operations and have more stringent constraints.
The approximate operations can be used to replace the multiplication operations in the error-tolerant systems which meet the constraints.

However, an arbitrary CNN can not always meet the constraints of approximate operation.
Therefore, certain transformations are applied to the general CNN to make them meet the constraints.
Then, the approximate operations proposed in the section~\ref{sec:approx} can replace the multiplication operations in the convolutional layers, thereby building CNNs with this approximate convolutional layers, named MinConvNets.
Section~\ref{sec:building} describes the methods to build and train MinConvNets.
Then, the proposed architecture is applied to common networks and data sets, and analyze the feasibility by the accuracy rate.
Section~\ref{sec:experiment} focuses on the experimental results of the approximate network.
Finally, the conclusion and the future work are discussed in  section~\ref{sec:conclusion}.

\vspace{-0.2cm}
\section{Approximate Operations to Multiplication}
\vspace{-0.15cm}
\label{sec:approx}
Similar operation and approximate operation to multiplication are proposed in this section.
The similar operation can replace multiplication if followed by a linear transformation, while the approximate operation based on similar operation can directly replace multiplication without transformation.

\subsection{Similar Operations}
\label{subsec:similar}
\begin{figure*}
    \centering
    \includegraphics[width=0.9\linewidth, trim=0 13cm 8cm 0, clip, page=1]{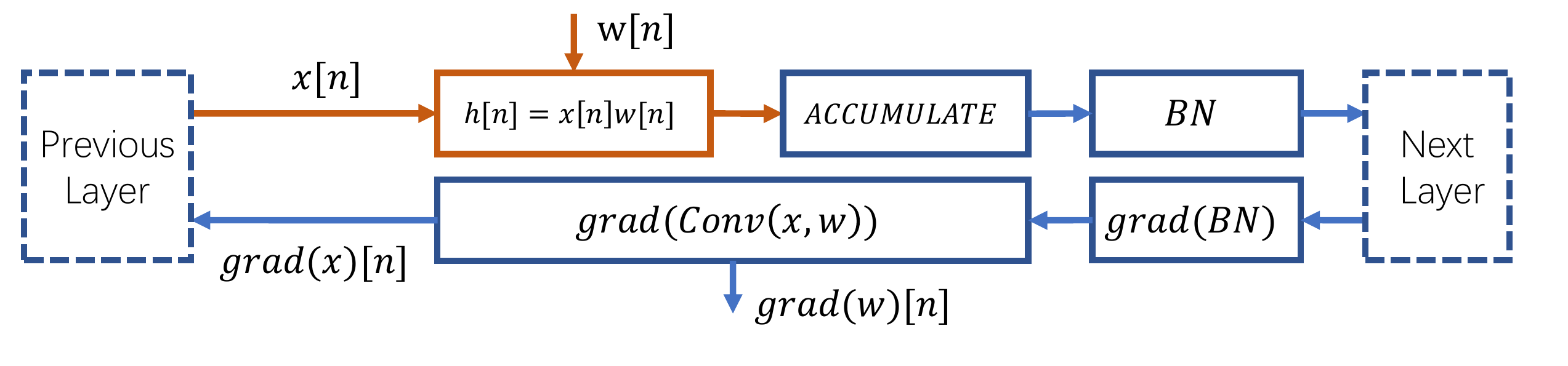}
    \vspace{-0.6cm}
    \caption{The convolutional layer is considered as a system.}
    \vspace{-0.5cm}
    \label{fig:conv_sys}
\end{figure*}
Taking a 2D spatial convolution convolution $Z = X \circledast  W$ in convolutional layers as an example, the elements in $Z$ are calculated as:
\vspace{-0.1cm}
\begin{equation}
    \vspace{-0.25cm}
    z_{u,v} =\sum\limits_{i=1}^{f_h} \sum\limits_{j=1}^{f_w} x_{u+i,v+j}\cdot w_{i,j}
    \label{eq:normal_mm}
\end{equation}
where $(u,v)$ shows the coordinate of pixels in the image, and $f_h, f_w$ are the height and width of convolutional kernel.
If the 2D arrays $x$ and $w$ are unfolded to 1D vectors according to their coordinate, Equation~\ref{eq:normal_mm} can be transformed into a multiply-accumulate operation as following:
\vspace{-0.15cm}
\begin{equation}
\vspace{-0.15cm}
    z_{u,v} =\sum\limits_{n=1}^{N} \boldsymbol{x}(n)\cdot \boldsymbol{w}(n)
    \label{eq:normal_mac}
\end{equation}
where $\boldsymbol{w}$ is the vector of $W$, $\boldsymbol{x}$ is the vector corresponding to the images pixels scanned by this filter at position $(u,v)$, and $N$ represents the length of the vector calculated as $N=f_w\cdot f_h$.
Equation~\ref{eq:normal_mac} can be seen as a system $H$ which takes two spatial signal $x$ and $w$,
then multiply them and generated the signal $h[n]=H(x[n],w[n])=x[n]\cdot w[n]$ orderly, followed by an accumulation operation.
From a perspective of system, this is a module of a convolutional neural network system, shown as Figure~\ref{fig:conv_sys}.

If there is another system $g[n]=G(x[n],w[n])$, the dependence of the output signals $h$ and $g$ with the same input signals is an indicator of the similarity between the systems $H$ and $G$.
The most familiar measure of dependence between two quantities is the Pearson product-moment correlation coefficient (PPMCC), commonly called simply "the correlation coefficient".
The correlation coefficient between two signals $h$ and $g$ is defined as follows:
\vspace{-0.2cm}
\begin{equation}
\vspace{-0.2cm}
\begin{array}{rl}
    \rho(h,g) &= \displaystyle{\frac{cov(h,g)}{\sqrt{var(h)var(g)}}}
\end{array}
\end{equation}
where $cov(h,g)$ is the covariance of $h$ and $g$, and $var(h)$ is the variance of $h$.
The correlation attempts to establish a line of best fit through two signals, and the correlation coefficient indicates the degree of the linear fit of the two signals~\cite{dowdy2011statistics}.
In other words, signals with strong correlation can generate approximate signals to each other by linear transformation.

\vspace{-0.1cm}
\begin{definition}[Similar Operations]
Let two systems take the same input signals, they are similar systems if they can generate strong correlation signals,
and the corresponding operations are similar operations.
\end{definition}
\vspace{-0.2cm}

For the system shown as Figure~\ref{fig:conv_sys}, the multiplication subsystem can be replaced by a similar system followed by a linear transformation,
to generate the approximate output.
Even though the approximate output brings errors, some works on few-bits quantization as~\cite{choi2018pact,choi2019accurate} have shown that CNNs are error-tolerant systems and the limited errors can be accepted.
It should be noted that in the forward propagation, the approximate output is considered as a result of exact operations with some errors, 
therefore it is still the exact operations that are used to calculate gradients in backward propagation.
The widely used training method SGD~\cite{bottou2010large} uses the approximate gradient to replace the real gradient of the forward propagation, 
which shows that the operations in forward propagation and the operations used to calculate gradient during backward propagation can be different.
Therefore, even though the real gradient of the forward propagation is different from the calculated gradient in the backward propagation in approximate convolution,
we believe that errors caused by similar operations have a limited impact on the training of CNNs.

Let $\mu_h$ and $\mu_g$ be the expected values of the signals $h[n]$ and $g[n]$, the covariance and variances for an input with a convolution kernel size of $N$ are calculated as follows:
\vspace{-0.18cm}
\begin{equation}
\vspace{-0.18cm}
    \left\{
    \begin{array}{rl}
    cov(h,g) &= \sum\limits_{n=1}^{N} (h[n]-\mu_h)(g[n]-\mu_g) \\
    var(h) &= \sum\limits_{n=1}^{N} (h[n]-\mu_h)(h[n]-\mu_h) 
    \end{array}
    \right.
\end{equation}
But what we need is to build approximate systems for arbitrary convolutions instead of for one determined convolution with fixed weights and images.
Therefore, $x$ and $w$ can be modelled as random variable with a probability distribution, hence $g$ and $h$ are random variable, too.
For commonly used image sets, the images are usually independent of each other.
If the images for each training iteration are randomly selected, 
the probability of images used in a new iteration has no relation to the weights trained in the previous iteration.
As well, the new images do not affect the probability distribution of the weights trained in the previous iteration, too.
Therefore, the weights and the images are considered as independent variables in each iteration.
Let $p_x(x)$ and $p_w(w)$ be the probability density functions of $x$ and $w$ respectively,
for the two given systems $h=H(x,w)$ and $g=G(x,w)$, the covariance and variances of $h$ and $g$ are calculated as:
\vspace{-0.15cm}
\begin{equation}
    \left\{
    \begin{array}{rl}
    cov(h,g) =& \int_{\xi} \int_{\eta}
        (H(\xi,\eta)-\mu_h)(G(\xi,\eta)-\mu_g) \\
        &\cdot p_x(\xi) p_w(\eta) d\xi d\eta \\
    \vspace{-0.3cm}
    &\\
    var(h,h) =& \int_{\xi} \int_{\eta}
        (H(\xi,\eta)-\mu_h)(H(\xi,\eta)-\mu_h) \\
        &\cdot p_x(\xi) p_w(\eta) d\xi d\eta
    \end{array}
    \right.
\end{equation}
where $\mu_h$ and $\mu_g$ is the expected values of $h$ and $g$.

\begin{table*}
    \centering
     \scalebox{0.75}{
    \begin{tabular}{c||c|c|c}
     Correlation with & min-selector &addition &  max-selector  \\
       $h[n] = |w[n]\cdot x[n]|$ 
     & 
        $
        g[n]=\left\{
        \begin{array}{rc}
        |w[n]| &\mbox{if } |w[n]| \leq |x[n]|\\
        |x[n]| &\mbox{if } |w[n]| > |x[n]|
        \end{array} \right.
        $
    & $g[n] = |w[n]|+|x[n]|$ 
    & 
        $
        g[n]=\left\{
        \begin{array}{rc}
        |x[n]| &\mbox{if } |w[n]| \leq |x[n]|\\
        |w[n]| &\mbox{if } |w[n]| > |x[n]|
        \end{array} \right.
        $ \\ \hline \hline
   $\left\{ 
   \begin{array}{rl}
        x\sim& N(0,1)  \\
        w\sim& N(0,1) 
   \end{array}
   \right.
   $    & 0.908  & 0.882 & 0.673 \\ \hline
   $\left\{ 
   \begin{array}{rl}
        x\sim& N(0,1)  \\
        w\sim& N(0,10) 
   \end{array}
   \right.
   $     & 0.692 & 0.683 & 0.624 \\ \hline
   $\left\{ 
   \begin{array}{rl}
        x\sim& N(0,10)  \\
        w\sim& N(0,1) 
   \end{array}
   \right.
   $     & 0.692 & 0.683 & 0.624 \\ \hline \hline
   $\left\{ 
   \begin{array}{rl}
        x\sim& U(0,1)  \\
        w\sim& U(0,1) 
   \end{array}
   \right.
   $     & 0.962 & 0.926 & 0.641 \\ \hline
   $\left\{ 
   \begin{array}{rl}
        x\sim& U(0,10)  \\
        w\sim& U(0,1) 
   \end{array}
   \right.
   $    & 0.716 & 0.717 & 0.655 \\ \hline
   $\left\{ 
   \begin{array}{rl}
        x\sim& U(0,1)  \\
        w\sim& U(0,10) 
   \end{array}
   \right.
   $   & 0.716 & 0.717 & 0.655
    \end{tabular}

    }
    \vspace{-0.2cm}
    \caption{The correlation coefficient of absolute values, where $N(\mu,\sigma)$ means a normal distribution with expected value $\mu$, and variance $\sigma^2$, and $U(a,b)$ means a uniform distribution in a closed interval $[a,b]$.}
    \label{tab:correlation}
    \vspace{-0.3cm}
\end{table*}


Three operators are used to compare the similarities with multiplication: min-selector, addition, and max-selector.
In order to maintain the same monotonicity, that magnitudes and signs are treated separately, 
i.e. all the operations take absolute values of operands only,
thereby avoiding negative numbers which make multiplications monotonically decreasing.
The signs can be calculated by the XNOR gate independently, as introduced in~\cite{rastegari2016xnor}.
According to different operators and their distributions, their similarities are shown in Table~\ref{tab:correlation}.
From this table, it can be seen that:
\begin{itemize}
\vspace{-0.15cm}
    \item The correlation coefficients are all greater than 0.5. In statistics, these signals can be called correlation signals. We believe that this correlation is mainly due to their monotonous increase.
    \item In case that $x$ and $w$ follow the probability distribution $N(0,1)$ and $U(0,1)$, both min-selector and addition show strong correlation, and min-selector is better than addition.
    Since the calculation is all for absolute value, the variance in normal distribution or interval in uniform distribution can be reflected by the average of the absolute value. 
    We note the above constraint that makes min-selector and multiplication strong correlation as $\mu_{|x|} = \mu_{|w|}$, which means that the expected value of the absolute values of $x$ and $w$ are equal.
    \item When the absolute values of $x$ and $w$ are more different, the correlation is weaker.
\vspace{-0.15cm}
\end{itemize}


%

The probability distribution of weights is still a subject under study~\cite{bellido1993backpropagation,gallagher1997weight,go1999analyzing,deutsch2019generative,ratzlaff2019hypergan}.
But normal and uniform distribution are commonly used initialization weight distributions~\cite{glorot2010understanding,he2015delving,he2016deep}.
Meanwhile, ~\cite{ahmed2018trained} shows that many weights may fit the \textit{t location-scale} distribution.
When $w$ follows the above distribution, $h$ and $g$ bring strong correlation with the constraint $\mu_{|x|} = \mu_{|w|}$.
In fact, the $\mu_{|x|} = \mu_{|w|}$ is not only a constraint mathematically or experimental, but it also needs to be maintained in the point of view of the convolutional layer in CNN:
\vspace{-0.1cm}
\begin{itemize}
\vspace{-0.1cm}
    \item If most of the values in  $w(n)$ are smaller than $x(n)$, min-selector will highly probably select the value in $w$, therefore the features in $x$ lost.
    \item On the contrary, if most of the values in $w$ are larger than values in $x(n)$, no matter what $w$ is, min-selector will highly probably select the value in $x$, so that $w$ can not distinguish the pattern in images.
    \vspace{-0.1cm}
\end{itemize}
In short, the constraint $\mu_{|x|} = \mu_{|w|}$ makes $w$ and $x$ comparable, which is a guarantee for $w$ extracting the information in $x$ by using min-selector.

\begin{figure*}
    \centering
    \includegraphics[width=0.95\linewidth, trim=0 16cm 11cm 0, clip, page=3]{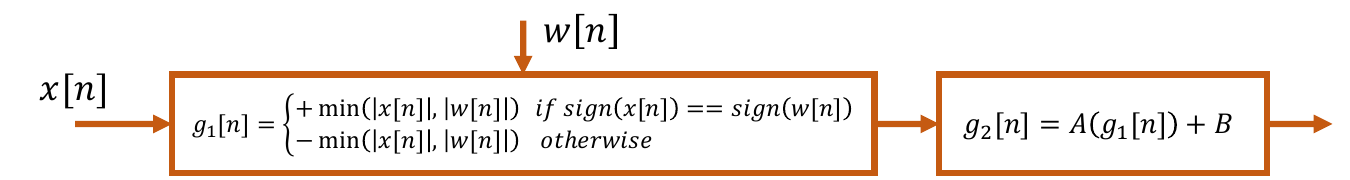}
    \vspace{-0.2cm}
    \caption{The approximate multiplication system constructed by the min-selector.}
    \vspace{-0.5cm}
    \label{fig:approx_mul}
\end{figure*}

Since multiplication and min-selector are similar operations with the constraint $\mu_{|x|} = \mu_{|w|}$,
a subsystem as shown in Figure~\ref{fig:approx_mul}  composed of min-selector and followed linear transformation can be built to replace the multiplication in the convolution system.
In this subsystem, the $g_1[n]$ takes the minimum absolute value of $x[n]$ and $w[n]$, and then keeps the sign consistent with the multiplication, that is strongly correlated with exact signal $h[n]$.
Then, a linear transformation $g_2$ is followed to generate the result approximate to $h[n]$.

\subsection{Approximate Operations}
\label{subsec:approx}
The similar signal is approximate to the exact signal if followed by a linear transformation as $g_2 = A(g_1)+B$.
However, the parameters $A$ and $B$ in this transformation are unknown.
To overcome this, we propose two directions:%
\begin{itemize}
    \item The parameters in linear transformation can be found through linear regression or machine learning.
    Methods for fitting linear models have been discussed in many works~\cite{merriman1877list,tibshirani1996regression,efron2004least}.
    But this means that in every iteration in the training stage, it needs to calculate the exact signal and approximate signal, then perform linear regression calculation to obtain $A$ and $B$, so this is a costly method. 
    We will not discuss the details.

    \item Supplementary constraints for similar operations should be proposed so that the linear transformation is not required in the approximate system.
    The correlation coefficient only reflects the consistency of the changing trend by the degree of linear fitness of two signals, regardless of their difference.
    In order to make the similar signal replace the exact signal without linear transformation, the difference between the signals should be as small as possible.
    Next, we discuss the supplementary constraint that makes the difference between two similar operations $g(n)$ and $h(n)$ small.
\end{itemize}

\begin{definition}[Approximate Operations]
Two similar systems are approximate if they have a small difference therefore they can replace each other in a fault-tolerant system, and the corresponding operations are approximate operations.
\end{definition}

We use the relative error between the approximate signal and the exact signal as a measure of their difference.
For a convolution with determined weights and images, the related error between $h[n]$ and $g[n]$ is calculated as following:
\vspace{-0.15cm}
\begin{equation}
\vspace{-0.15cm}
L(h,g) =\sum\limits_n \displaystyle{\left|\frac{g[n]]-h[n]}{h[n]}\right|}
\end{equation}

For arbitrary convolution systems, it is needed to consider the probability distribution of the input signals.
Let $p_x(x)$ and $p_w(w)$ be the probability density function of $x$ and $w$ respectively,
the difference between signals $h=H(x,w)$ and $g=G(x,w)$ is calculated as :
\vspace{-0.1cm}
\begin{equation}
    \begin{array}{rl}
    L(h,g) =& \int_{\xi} \int_{\eta}
        \displaystyle{\left|\frac{H(\xi,\eta)-G(\xi,\eta)}{H(\xi,\eta)}\right|}
        \cdot p_x(\xi) p_w(\eta) d\xi d\eta
    \end{array}
    \label{eq:relative_error}
\end{equation}

The supplementary constraints need to make $L(h,g)$ as small as possible, where $h$ and $g$ are similar signals.
We take the similar signal $h=H(x,w)=|w\cdot x|$ and $g=G(x,w)= min(|w|,|x|)$ as example, where $w$ and $x$ follow normal distributions. 
In order to ensure the constraint of similar operation $\mu_{|x|} = \mu_{|w|}$, we set that $x$ and $w$ have the same averages and related small variance.
Therefore, $w,x\sim N(k,v)$ where $N(k,v)$ means the normal distribution with average $k$ and variance $v$.
With these conditions, $L(h,g)$ is a function about $(k,v)$, 
and the task of finding the supplementary constraint is transformed into an optimization problem, that is, finding the values of $(k,v)$ which minimize $L(k,v)$.

\begin{figure*}
   	\centering
    \subfloat[$L(k)$ with different $v$.]
   	{
 	    \includegraphics[width=0.32\linewidth, trim=0 0 2cm 0, clip, page=7]{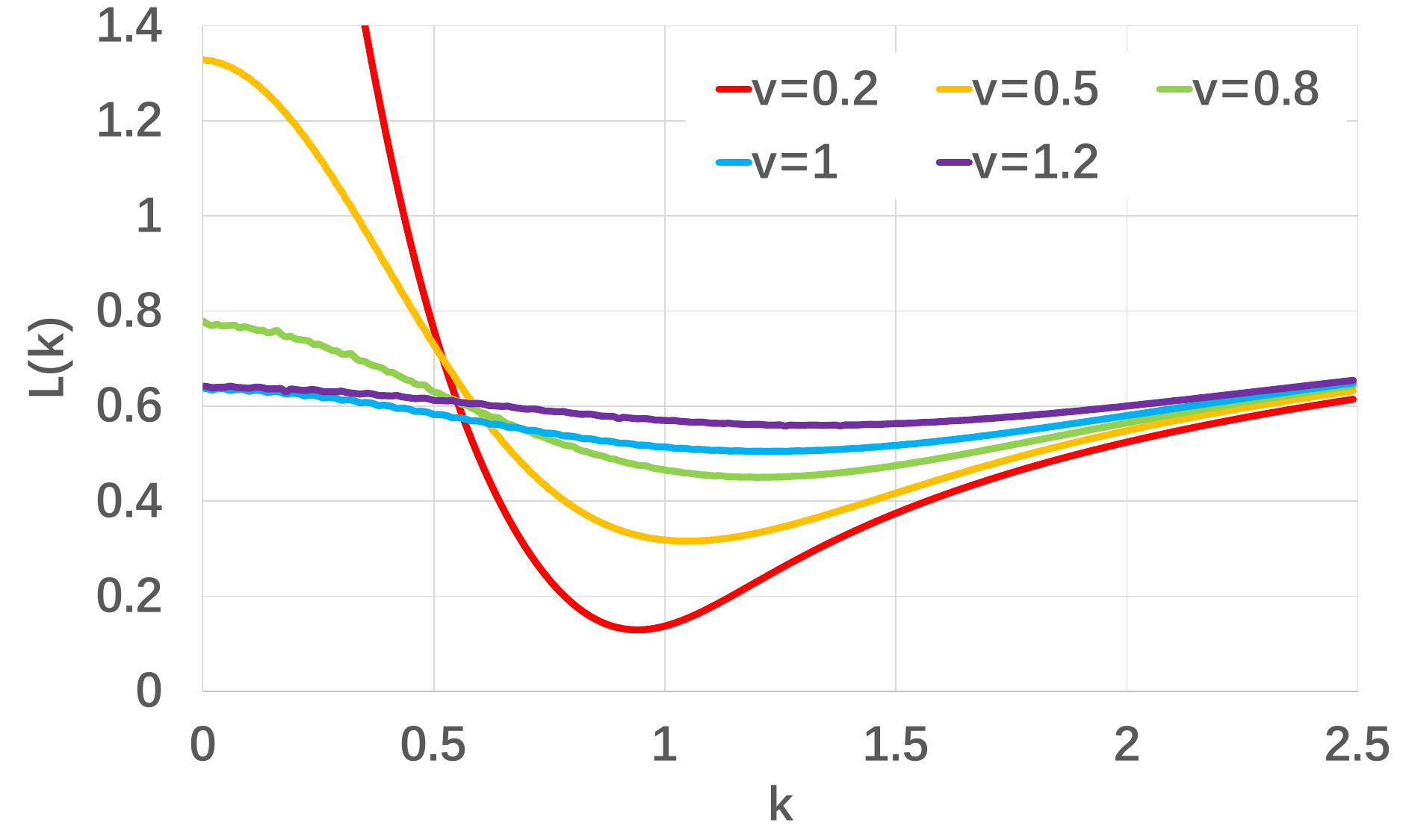}
        \label{fig:Lk}
    }
   	\hfill
    \subfloat[$k(v)$ with minimum $L(k)$.]
   	{
 	    \includegraphics[width=0.32\linewidth, trim=0 0 2cm 0, clip, page=8]{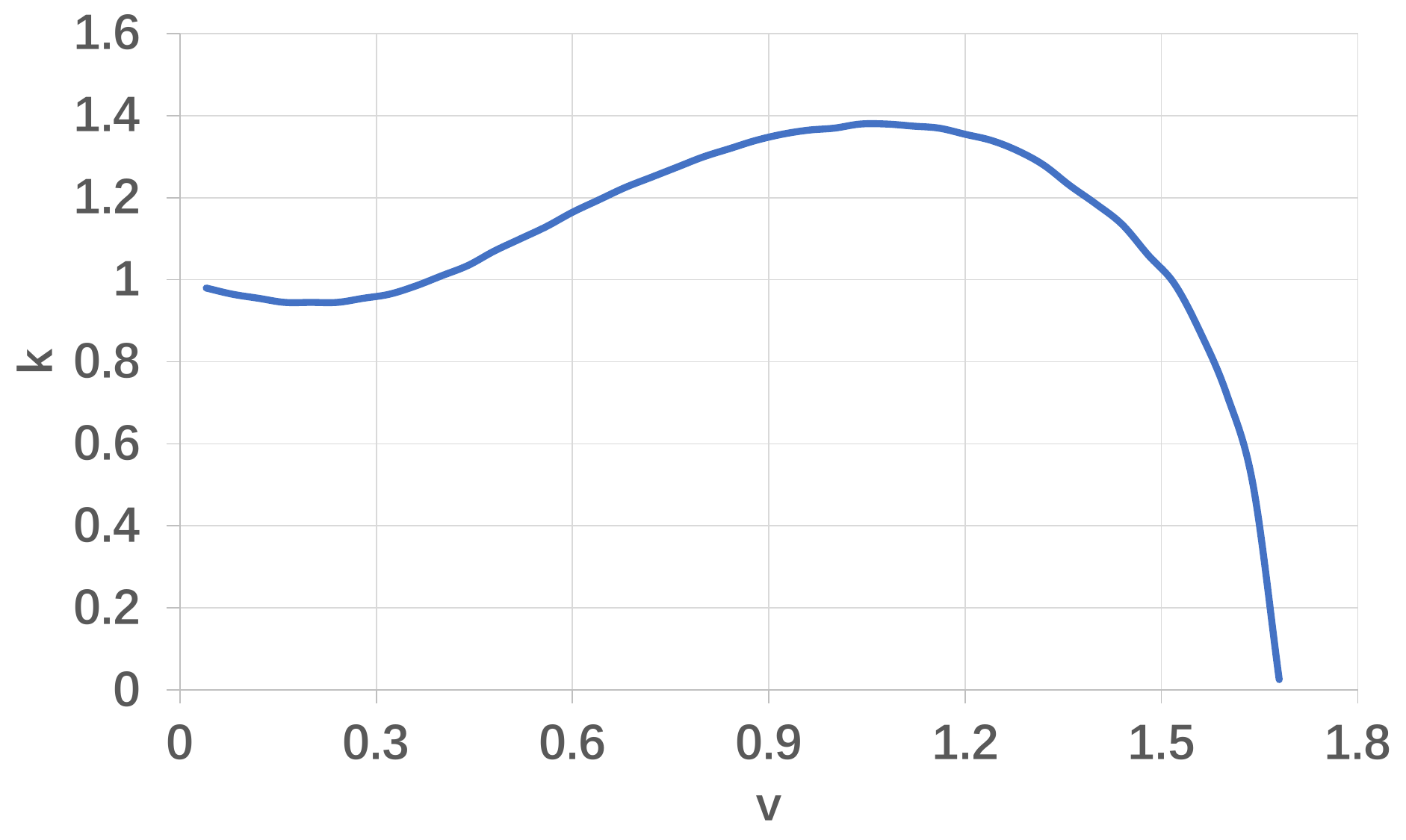}
        \label{fig:kv}
    }
   	\hfill
    \subfloat[$L(v)$ with $k=0$.]
   	{
 	    \includegraphics[width=0.32\linewidth, trim=0 0 2cm 0, clip, page=9]{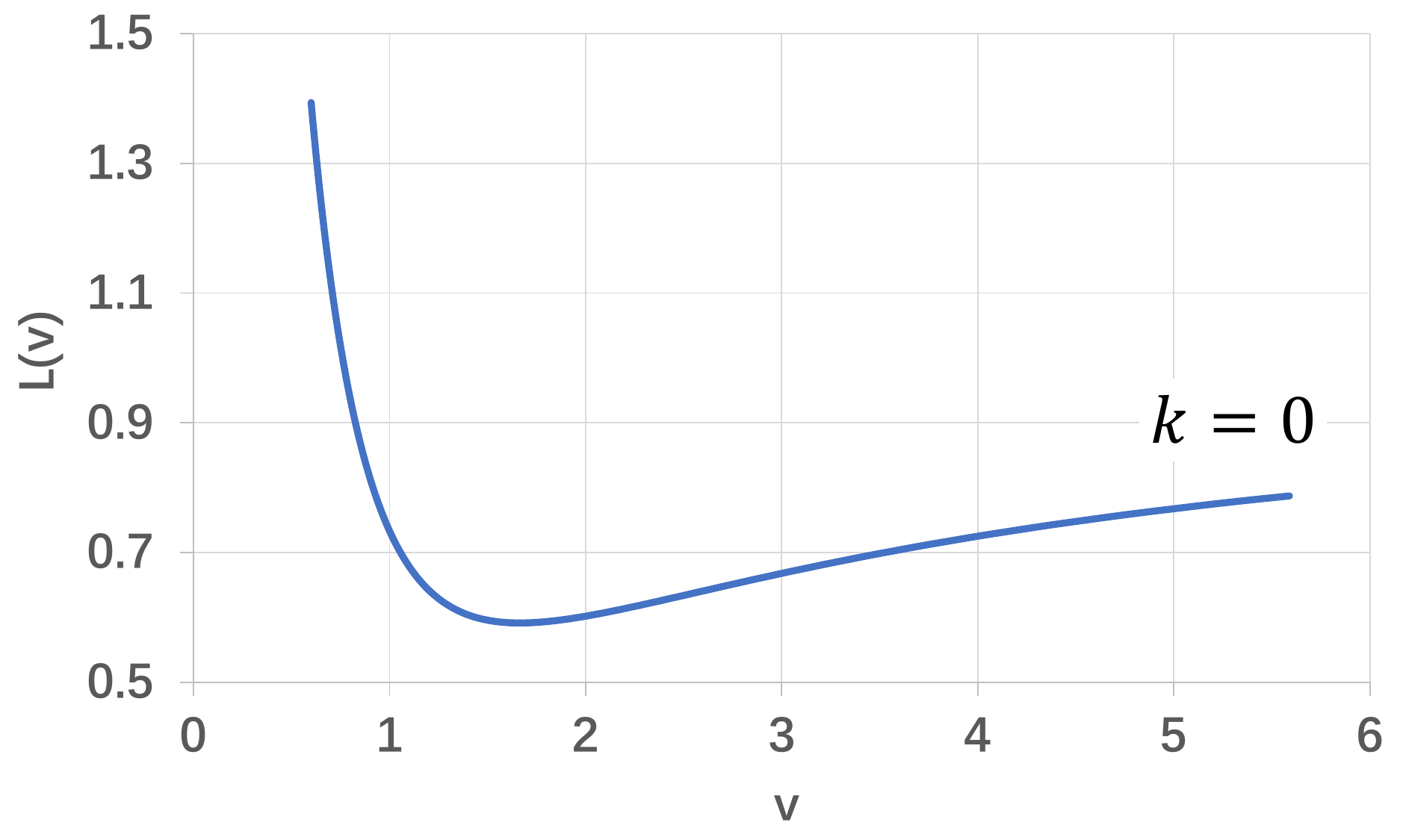}
        \label{fig:Lv}
    }
   	\caption{The figures about $L,k,v$.}
   	\label{fig:related_error}
   	\vspace{-0.4cm}
 \end{figure*}

When $\xi=0$ or $\eta=0$ there are singularities in Equation~\ref{eq:relative_error} because of the division by $H(\xi,\eta)=0$.
Therefore, the points which make $H(\xi,\eta) = 0$ are break-points and are removed from integration interval.
In fact, if $\xi$ or $\eta$ is zero, $H(\xi,\eta)=H(\xi,\eta) = 0$, so  there is no error between the two signals. 
Figure~\ref{fig:Lk} shows the the function $L(k)$ with varied variance $v$.
It can be seen that $L$ achieves a minimum value when $k$ is around $1$.
Figure~\ref{fig:kv} shows the $k$ that makes $L(k,v)$ minimum with different variances $v$.
It can be seen that the value of $k$ is always concentrated around $1$ if the variance is less than $1.5$.
In other words, a quantitative relationship can be obtained: the more values of the operands are concentrated to $1$ (or $-1$ for negative), the more approximate the signals $h$ and $g$ are.
In fact, with the application of the batch normalization, the inputs of hidden layers in the CNNs have been standardized, so that we assume $x$ distributed as $x\sim N(0,1)$.
Since the variance $v$ can be changed by simply multiplying by a constant, we study the effect of the variance $v$ on $L$.
Figure~\ref{fig:Lv} shows how the value of $L$ changes with variance when $k=0$.
It can be seen that when $v=1.57$, $L$ takes the minimum value.
For the normal distribution, the average value of the absolute value at $v=1.57$ can be calculated by its variance, $\mu_{|x|} = 1.25$, that is not far from $1$.

Based on the above analysis, we propose a constraint of approximation that the absolute value of operands should be concentrated around $1$. 
Taking into account the constraint of similarity, we mark the constraint with which min-selector is approximate to multiplication as $\mu_{|x|}=\mu_{|w|} = 1$,
i.e., the values of the two input parameters are as close as possible, and concentrated to $1$ or $-1$.
This is not an exact mathematical condition, but it is the main guide for building a multiplication-less CNN with min-selectors in the next section.


\vspace{-0.2cm}
\section{Building Approximate Networks}
\vspace{-0.1cm}
\label{sec:building}
In section~\ref{sec:approx}, we propose min-selector which is an approximate operation to multiplication with the constraint $\mu_{|x|}=\mu_{|w|} = 1$.
However, the multiplication in arbitrary neural networks does not always meet the constraint.
Therefore, in this section, we simply transform the convolutional layer to make it meet the constraints, and then use approximate operations to construct the convolutional layer without multiplication.

\subsection{Building the Approximate Convolution}
\vspace{-0.15cm}
\label{subsec:build_approx}
Next, we proposed the transformation to make an arbitrary multiplication $z=x\cdot w$ meet the approximate constraint $\mu_{|x|}=\mu_{|w|} = 1$.
To keep the same monotonicity between multiplication and min-selector, the value and sign of output results are processed separately.

The first step is to set the expected value of the absolute value of the operands to $1$:
\vspace{-0.2cm}
\begin{equation}
\vspace{-0.15cm}
    \frac{|z|}{\mu_{|x|} \mu_{|w|}} = \frac{|x|}{\mu_{|x|}}\cdot \frac{|w|}{\mu_{|w|}}
\end{equation}
where $\mu_{|x|}$ and $\mu_{|w|}$ are the expected value of $|x|$ and $|w|$.
At this time, the two operands for multiplication are $\frac{|x|}{\mu_{|x|}}$ and $\frac{|w|}{\mu_{|w|}}$, and their expected values are both $1$. 
Then, the min-selector can be used to replace the multiplication to generate an approximate value, as shown below.
\vspace{-0.2cm}
\begin{equation}
\vspace{-0.15cm}
    \frac{|z|}{\mu_{|x|} \mu_{|w|}} \approx min(\frac{|x|}{\mu_{|x|}}, \frac{|w|}{\mu_{|w|}})
    \label{eq:build_eq}
\end{equation}

According to the nature of min-selector and multiplication, the Equation~\ref{eq:build_eq} is transformed to obtain the following equation:
\vspace{-0.2cm}
\begin{equation}
\vspace{-0.15cm}
    \begin{array}{rl}
    |z| & \approx \mu_{|x|}\cdot \mu_{|w|} \cdot min(\displaystyle{\frac{|x|}{\mu_{|x|}}}
    , \displaystyle{\frac{|w|}{\mu_{|w|}}}) \\
    &\\
      &= \mu_{|x|}\cdot \mu_{|w|} \cdot ( \displaystyle{\frac{1}{\mu_{|x|}}} \cdot min(|x|
    , \displaystyle{\frac{\mu_{|x|}}{\mu_{|w|}}}\cdot |w|))\\
    &\\
      &= \mu_{|w|} \cdot min(|x|
    , \displaystyle{\frac{\mu_{|x|}}{\mu_{|w|}}}\cdot |w|)\\
    \end{array}
\end{equation}

Since the sign of min-selector should be consistent with multiplication, the approximate calculation for the signed operands is: 
\vspace{-0.2cm}
\begin{equation}
\vspace{-0.15cm}
    \begin{array}{cc}
    z \approx   &\mu_{|w|} \cdot smin(x
    , \displaystyle{\frac{\mu_{|x|}}{\mu_{|w|}}}\cdot w)
    \end{array}
\end{equation}
where $smin(a,b)$ is for signed operands and calculated as Table~\ref{tab:smin}.
It can be seen that $smin(a,b)$ is calculated by comparing the values and signs of operands separately.
And there are only four possible outputs according to the comparison, so that can be implemented in a 4-to-1 multiplexer with comparators, which is friendly to some platforms such as FPGA~\cite{deschamps2012guide}.

\begin{table}
    \centering
        \begin{tabular}{c||c|c}
         $smin(a,b) $& $|a|\leq |b|$ & $|a|>|b|$ \\ \hline \hline
         $sign(a)=sign(b)$ & $|a|$ & $|b|$\\  \hline
         $sign(a)\neq sign(b)$ & $-|a|$ & $-|b|$
    \end{tabular}

    \caption{The operation $smin(a,b)$.}
    \vspace{-0.6cm}
    \label{tab:smin}
\end{table}

For a well-trained neural network, the weights $W$ and $\mu_{|w|}$ are known.
Inspired by~\cite{ioffe2015batch} which uses moving averages in batch normalization to fix the averages and variances during inference, the moving $\mu_{|x|}$ obtained through training is used to fix $\mu_{|x|}$ as a constant during inference.
Therefore, for a well-trained network, $\mu_{|x|}$ and $\mu_{|w|}$ are known, so we can generate new weights $\widetilde{W}$ where the elements are calculated as:
\begin{equation}
    \widetilde{w} =  \displaystyle{\frac{\mu_{|x|}}{\mu_{|w|}}}\cdot w
\end{equation}
which can be directly used for comparison with the inputs without recalculation during the inference.

Then the convolution in Equation~\ref{eq:normal_mm} can be calculated by approximate operations as follows:
\vspace{-0.2cm}
\begin{equation}
\vspace{-0.15cm}
    \begin{array}{rl}
        z_{u,v} &\approx \sum\limits_{i=1}^{f_h} \sum\limits_{j=1}^{f_w} ( \mu_{|w|} \cdot smin(x_{u+i,v+j}, \widetilde{w}_{i,j})) \\
        &= \mu_{|w|}\cdot \sum\limits_{i=1}^{f_h} \sum\limits_{j=1}^{f_w} smin(x_{u+i,v+j}, \widetilde{w}_{i,j}) 
    \end{array}
    \label{eq:approx_mm}
\end{equation}


Although there is still $1$ multiplication operation in Equation~\ref{eq:approx_mm}, it is much fewer than the $f_h\cdot f_w$ multiplication operations in Equation~\ref{eq:normal_mm}.
Since many convolutional layers are followed by the bias layer or the batch normalization layer, the multiplication in Equation~\ref{eq:approx_mm} can be integrated into these following layers, thereby constructing an approximate convolutional layer that does not require multiplication.


\subsection{Training Approximate Convolutional Layers}
Based on the approximate convolution constructed in section~\ref{subsec:build_approx}, we can build an approximate convolution layer.
We call the neural network composed of the approximate convolutional layers MinConvNets.
Algorithm~\ref{algo:approx_layer} shows how to train a MinConvNets.
\begin{algorithm}
\hspace*{\algorithmicindent} \textbf{Input}: A batch of inputs and targets $(X,Y)$, cost function $cost(Y,\hat{Y})$, current weight $W^t$ and current learning rate $\eta^t$. 

\hspace*{\algorithmicindent} \textbf{Output}: updated weight $W^{t+1}$ and updated learning rate $\eta^{t+1}$.
\setstretch{1.6} 
\begin{algorithmic}[1]
  \STATE //\textbf{Forward propagation}
  \STATE {\textbf{for} $l=1$ \textbf{to} $L$ \textbf{do}}
  \STATE \quad$\mu^r_{|x_{l}|} = \gamma \cdot \mu^r_{|x_{l}|} + (1-\gamma)\cdot  \mu_{|x_{l}|}$
  \STATE \quad$X_{l}^c = clip(X^t_{l}, 2 \mu_{|x_{l}|})$
  \STATE \quad\textbf{for} $k^{th}$ filter in $l^{th}$ layer \textbf{do}
  \STATE \qquad$W_{lk}^c = clip(W^t_{lk}, 2 \mu_{|w_{lk}|})$
  \STATE \qquad$\widetilde{W}_{lk} = \displaystyle{\frac{\mu_{|x_{lk}|}}{\mu_{|w_{lk}|}}} \cdot W_{lk}^c$
  \STATE \quad$\hat{Y}_l=$\textbf{ApproxForward}$(X_l^c,\widetilde{W}_l)$

  \STATE //\textbf{Backward propagation}
  \STATE $C = cost(Y, \hat{Y}_L)$
  \STATE {\textbf{for} $l=L$ \textbf{to} $1$ \textbf{do}}
  \STATE \quad $\frac{\partial{C}}{\partial{X^c_l}},\frac{\partial{C}}{\partial{W^c_l}}=$\textbf{ExactBackward}$(\frac{\partial{C}}{\partial{\hat{Y}_l}},W^c_l,X_l^c)$
  \STATE \quad $\frac{\partial{C}}{\partial{W^t}}=\frac{\partial{C}}{\partial{W^c_l}} \cdot \frac{\partial{W^c_l}}{\partial{W^t_l}}$
  \STATE \quad $\frac{\partial{C}}{\partial{\hat{Y}_{l-1}}}=\frac{\partial{C}}{\partial{X^t}}=\frac{\partial{C}}{\partial{X^c_l}} \cdot \frac{\partial{X^c_l}}{\partial{X^t_l}}$
  
  \STATE //\textbf{Update Patameters}
  \STATE $W^{t+1}=$\textbf{UpdateParamters}$(W^t,\frac{\partial{C}}{\partial{{W^t}}},\eta^t)$
  \STATE $n^{t+1}=$\textbf{UpdateLearningrate}$(\eta^t,t)$
\end{algorithmic}
\caption{Training an L-layers CNN with approximate convolutions.}
\label{algo:approx_layer}
\end{algorithm}

    At the beginning of forward propagation, the moving expected value is used to record the average of the absolute value $\mu^r_{|x|}$, where $\gamma$ is the momentum of moving value.
    While $\mu_{|x|}$ is used during training, a well trained $\mu^r_{|x|}$ can be used to replace $\mu_{|x|}$ during the inference phase to avoiding calculations.
 
    $X$ and $W$ are clipped by $clip$ function, and
    $clip(X,\alpha)$ means $\forall x\in X$ the calculation following is applied:
    \begin{equation}
        clip(x,\alpha) =
        \left\{
        \begin{array}{rl}
            \alpha & \mbox{if } x>\alpha \\
            -\alpha & \mbox{if } x<-\alpha \\
            x & \mbox{otherwise}
        \end{array}
        \right.
    \end{equation}
    Correspondingly, the gradient is calculated as follows:
    \begin{equation}
        grad(clip(x,\alpha)) =
        \left\{
        \begin{array}{rl}
            0 & \mbox{if } x>\alpha \mbox{ or } x<-\alpha \\
            1 & \mbox{otherwise}
        \end{array}
        \right.
    \end{equation}
    The main purpose of the $clip$ function is to avoid excessive variance caused by excessive values which may reduce the approximation of operations.
    For a well-trained network, the pre-clipped weights can be used for the inference stage. And the inputs do not need to be clipped during the inference stage, because with the clipped weights, too large input values will not be selected by min-selector. 
    Therefore, the $clip$ function is a method to ensure that the network converges, but it does not increase the amount of calculation during the inference stage.
    
    $ApproxForward$ is a standard forward propagation, except that the matrix multiplication calculated as shown in Equation~\ref{eq:approx_mm}.
    The result of the convolution needs to be multiplied by the constant $\mu_{|w|}$. 
    As mentioned in section~\ref{subsec:build_approx}, this operation can be done directly or integrated into subsequent layers.
    The $ApproxForward$ uses $\{ X^c_l , \widetilde{W}_l\}$ to calculate the approximate convolution of $\{X^c_l , W^c_l\}$. 
    Therefore, it still calculates the gradient of an exact convolution for backward propagation, i.e.,  $\{X^c_l , W^c_l\}$ are used as inputs instead of $\{ X^c_l , \widetilde{W}_l\}$.
    
    Any update rules (e.g., SGD or ADAM) and learning rate scheduling functions can be applied to update the parameters and learning rate at the end of the algorithm.

\vspace{-0.2cm}
\section{Experiments}
\vspace{-0.1cm}
\label{sec:experiment}
In order to compare our multiply-less architecture with standard results, we build different networks and apply them to different data sets.

\subsection{Networks and Image Sets}
\vspace{-0.1cm}
\begin{table}[]
    \centering
    \scalebox{0.9}{
            \begin{tabular}{cc}
        \hline Layer & Size \\ \hline
        Input image & 28x28x3 \\
        Conv.+ leaky ReLU & 5x5x32 \\
        MaxPool. & 2x2 \\
        Conv.+ leaky ReLU & 5x5x64 \\
        Fully Connected+ReLU & 1024 \\
        drop out & 50\% \\
        Fully Connected & 10 \\\hline
    \end{tabular}

    }
    \caption{LeNet network.}
    \label{tab:lenet}
\end{table}

\begin{table}[]
    \centering
    \scalebox{0.9}{
            \begin{tabular}{cc}
        \hline Layer & Size \\ \hline
        Input image & 28x28x3 \\
        Conv.+ leaky ReLU & 3x3x32 \\
        MaxPool. & 2x2 \\
        Conv.+ leaky ReLU & 1x1x16 \\
        Conv.+ leaky ReLU & 3x3x64 \\
        MaxPool. & 2x2 \\
        Conv.+ leaky ReLU & 1x1x32 \\
        Conv.+ leaky ReLU & 3x3x128 \\
        Conv.+ leaky ReLU & 1x1x64 \\
        Fully Connected+ReLU & 1024 \\
        drop out & 50\% \\
        Fully Connected & 10 \\\hline
    \end{tabular}

    }
    \caption{mini\_cifar network.}
    \label{tab:mini_cifar}
\end{table}
\begin{table*}[]
    \centering
    \scalebox{1}{


\begin{tabular}{cl|c|c|c}
    Architecture && LeNet-MNIST & LeNet-Cifar10 & mini\_cifar-Cifar10 \\ \hline
    Standard Network &                & 99.06\% & 75.26\% & 77.30\% \\ \hline
     \multirow{3}*{Approximate}       & 170 epoch         & 98.42\% &         &         \\
     ~              & 512 epoch     &         & 64.18\% & 71.46\% \\
     ~          & 2048 epoch        &         & 65.54\% & 72.89\% \\ \hline
    \multirow{2}*{Transfer Learning} & 512 epoch   &         & 74.92\% & 77.01\% \\
    ~ &1024 epoch  &         & 75.10\% & 77.26\%
\end{tabular}



    }
    \caption{Accuracy for different architectures and image sets.}
    \label{tab:accuracy}
    \vspace{-0.25cm}
\end{table*}

LeNet and mini\_cifar are used to build approximate networks.
LeNet shown in Table~\ref{tab:lenet} is a small convolutional network for image classification.
In order to measure the performance for negative operands, leaky ReLU~\cite{redmon2017yolo9000} instead of ReLU is used in convolutional layers as the activation function.
The mini\_cifar shown in Table~\ref{tab:mini_cifar} is a network deeper than LeNet.
There are $6$ convolutional layers, but three of them are 1x1 convolutional layers, mainly used to adjust the size of the network.

The image sets used were: MNIST and Cifar10.
MNIST~\cite{deng2012mnist} is a images-set of handwritten digits with 60,000 examples in the training set and  10,000 examples in the test set. On the other hand, 
Cifar10~\cite{krizhevsky2009learning} is a data-set of 60000 colored images in 10 classes with 50000 training images and 10000 test images.

\subsection{Approximate Networks}
\vspace{-0.1cm}
\begin{figure}
    \centering
 	    \includegraphics[width=0.9\linewidth, trim=0 0 2cm 0, clip, page=14]{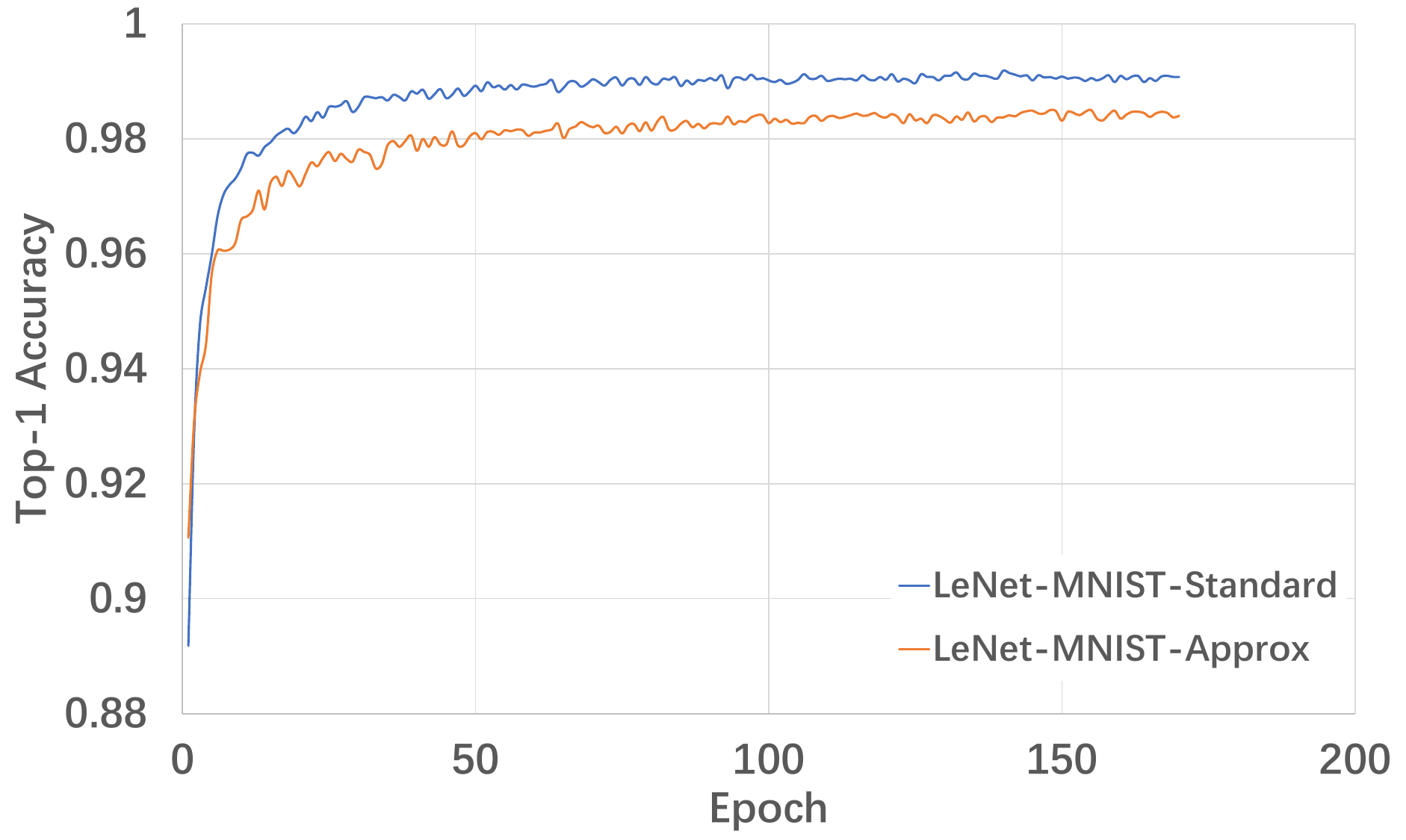}
    \caption{Top-1 accuracy of LeNet applied to MNIST.}
    \label{fig:lenet_mnist}
    \vspace{-0.2cm}
\end{figure}
\begin{figure}
    \centering
 	    \includegraphics[width=0.9\linewidth, trim=0 0 2cm 0, clip, page=15]{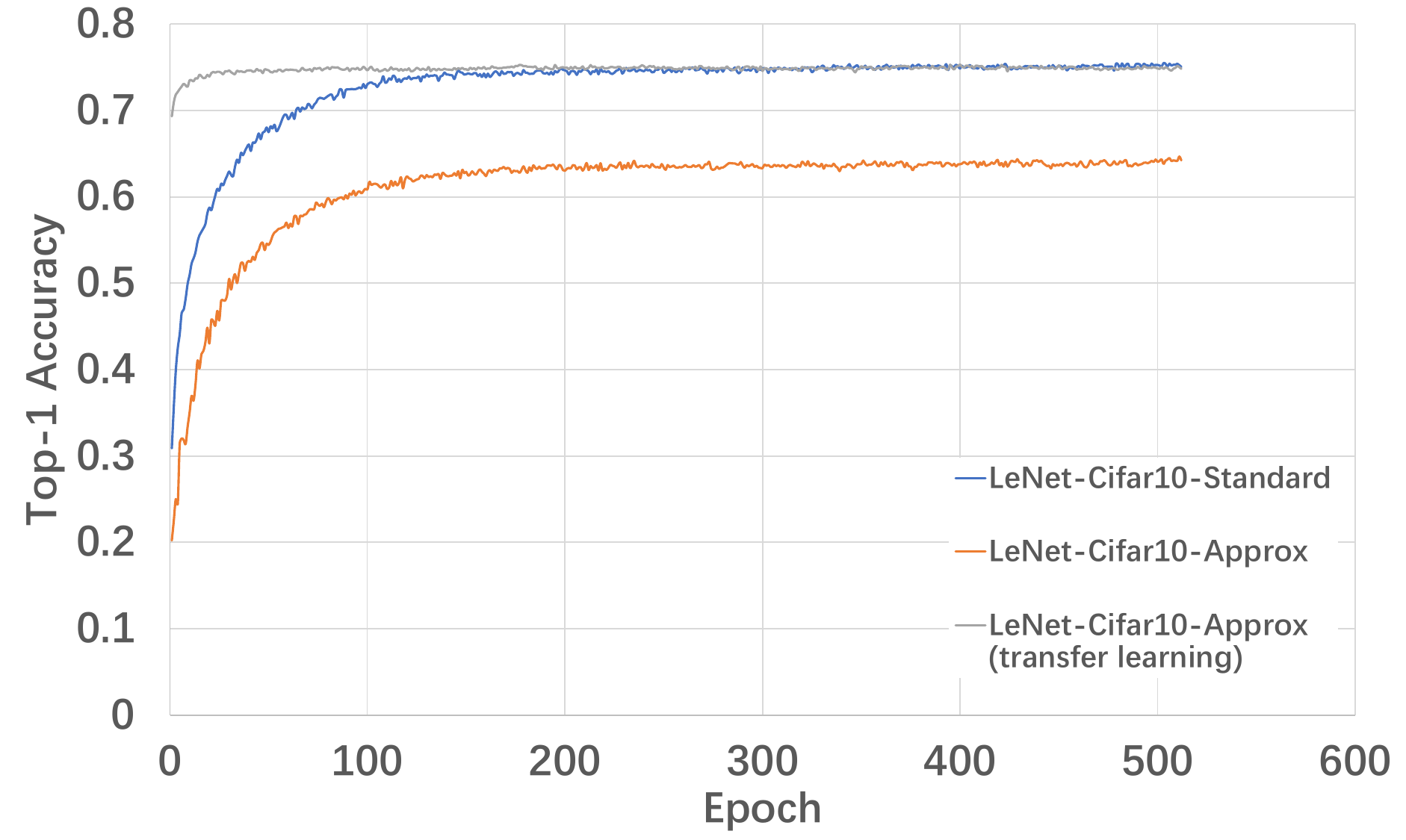}
    \caption{Top-1 accuracy of LeNet applied to Cifar10.}
    \label{fig:lenet_cifar10}
    \vspace{-0.5cm}
\end{figure}
\begin{figure}
    \centering
 	    \includegraphics[width=0.9\linewidth, trim=0 0 2cm 0, clip, page=16]{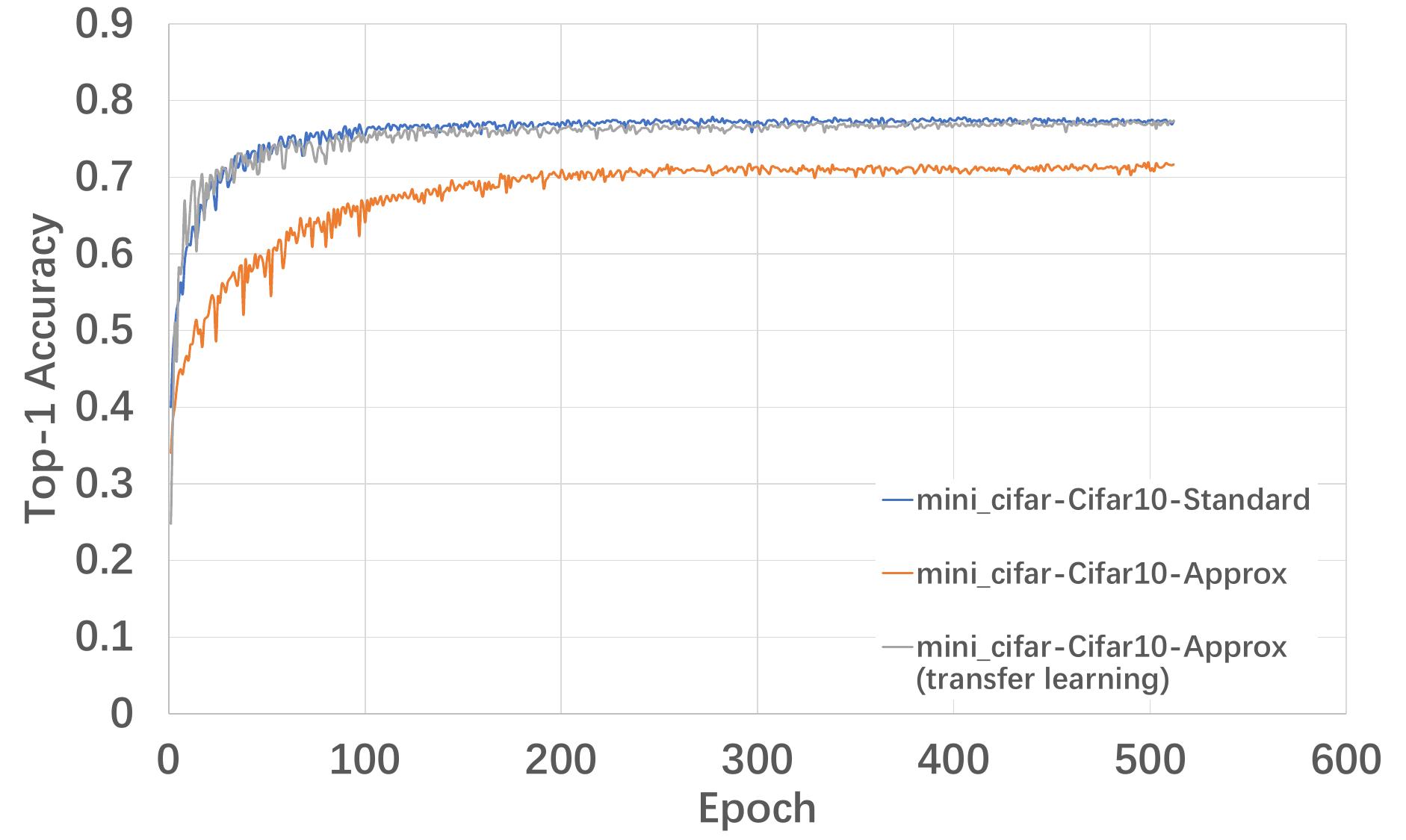}
    \caption{Top-1 accuracy of mini-cifar applied to Cifar10.}
    \label{fig:mini_cifar10}
    \vspace{-0.5cm}
\end{figure}

The accuracy of different networks and image sets are shown in Table~\ref{tab:accuracy}.
In order to evaluate the training speed easily, 
we recorded the accuracy increasing along with the training epoch, 
as shown in Figure~\ref{fig:lenet_mnist}, Figure~\ref{fig:lenet_cifar10}, and Figure~\ref{fig:mini_cifar10}.

Since MNIST is a relatively simple database, it is usually used to test whether the neural networks work. 
Although its accuracy is reduced by 0.65\% compared with the exact LeNet, the approximate LeNet achieves good accuracy.
It proves that the approximate network can converge and be used for image classification.

Compared with MNIST, Cifar10 is more complex.
Two networks are applied to Cifar10, with exact or approximate convolution.
The accuracies of different networks are shown in Table~\ref{tab:accuracy}.
It can be seen that for more complex data sets, approximate networks have brought errors compared with exact networks.
Figure~\ref{fig:lenet_cifar10} and Figure~\ref{fig:mini_cifar10} show the accuracy along with training.
After 512 epoch, the accuracies of standard networks have already stopped increasing, while the approximate networks are still slowly rising.
On the one hand, this reflects the slower training of the approximate network. 
On the other hand, it also shows that the accuracy of the approximate network can be better with more training.
In fact, we continue training the approximate networks until the 2048 epoch, in which the increase is much slower, the results are shown in Table~\ref{tab:accuracy}.
Continued training brings a related better result, but it did not change our conclusion, that is, 
compared with the exact network, the training of approximate network is slower, 
and depending on the data set and network structure, the approximate network will bring different degrees of error.

\vspace{-0.05cm}
\subsection{Transfer Learning}
\vspace{-0.05cm}
Since MinConvNets are approximate to exact networks, 
we believe that the weights which have been well-trained on the exact networks 
can accelerate the convergence of the approximate network. 
Therefore, the transfer learning method is applied to train the approximate network.

The weights after 512 epoch trained in the standard network are used as the initial weights of the approximate network.
Then the top-1 accuracy along with the training epochs are shown in Figure~\ref{fig:lenet_cifar10} and Figure~\ref{fig:mini_cifar10}.
It can be seen from these figures that transfer learning can speed up the training of approximate networks.
After 512 epoch training, the accuracy levels of the approximate network are almost the same as the exact networks.
In addition, as shown in Table~\ref{tab:accuracy}, with more training, such as 1024 epochs, the accuracy is closer to the standard network.
In other words, there is negligible loss of accuracy in the approximate CNNs.

\vspace{-0.2cm}                         
\section{Conclusion}
\vspace{-0.15cm}
\label{sec:conclusion}
In order to speed up the CNNs, we reinterpret CNNs from the perspective of signals and systems.
and propose a new structure named MinConvNets.
In the convolutional layers of MinConvNets, the multiplication is approximated by an operation based on a minimum comparator, which is easier to be implemented or faster to be calculated.
MinConvNets bring negligible loss of accuracy in the benchmark test.

In addition, the MinConvNets have shown that operations with errors can also generate good global inference results for CNNs, so that multiplication can be replaced by some simpler operation.
This work shows that other operations can also extract features from images, and provides a new research direction for the acceleration technologies for neural networks.

This work is still in progress and various improvements are parts of future research.
First, the main contribution of this work is to propose a new architecture, but do not measure the runtime performance.
The proposed approximate operation can be implemented in different platforms, and it is necessary to measure the runtime performance on these platforms.
Next, other approximate calculators can be applied.
Some other more valuable approximate calculators may bring greater acceleration and structural optimization.
Furthermore, MinConvNets are still based on original neural networks, such as using the same training methods as original ones.
Since other operations can also extract features, it is possible to get out of the original framework and propose a more light image processing system in the future.


\newpage
{\small
\bibliographystyle{ieee_fullname}
\bibliography{main}

\begin{thebibliography}{10}\itemsep=-1pt

\bibitem{ahmed2018trained}
Muhammad Atta~Othman Ahmed.
\newblock Trained neural networks ensembles weight connections analysis.
\newblock In {\em International Conference on Advanced Machine Learning
  Technologies and Applications}, pages 242--251. Springer, 2018.

\bibitem{anwar2017structured}
Sajid Anwar, Kyuyeon Hwang, and Wonyong Sung.
\newblock Structured pruning of deep convolutional neural networks.
\newblock {\em ACM Journal on Emerging Technologies in Computing Systems
  (JETC)}, 13(3):1--18, 2017.

\bibitem{bellido1993backpropagation}
I Bellido and Emile Fiesler.
\newblock Do backpropagation trained neural networks have normal weight
  distributions?
\newblock In {\em International Conference on Artificial Neural Networks},
  pages 772--775. Springer, 1993.

\bibitem{bottou2010large}
L{\'e}on Bottou.
\newblock Large-scale machine learning with stochastic gradient descent.
\newblock In {\em Proceedings of COMPSTAT'2010}, pages 177--186. Springer,
  2010.

\bibitem{diannao}
Tianshi Chen, Zidong Du, Ninghui Sun, Jia Wang, Chengyong Wu, Yunji Chen, and
  Olivier Temam.
\newblock Diannao: A small-footprint high-throughput accelerator for ubiquitous
  machine-learning.
\newblock In {\em Proceedings of the 19th International Conference on
  Architectural Support for Programming Languages and Operating Systems},
  ASPLOS '14, pages 269--284, New York, NY, USA, 2014. ACM.

\bibitem{choi2019accurate}
Jungwook Choi, Swagath Venkataramani, Vijayalakshmi Srinivasan, Kailash
  Gopalakrishnan, Zhuo Wang, and Pierce Chuang.
\newblock Accurate and efficient 2-bit quantized neural networks.
\newblock In {\em Proceedings of the 2nd SysML Conference}, volume 2019, 2019.

\bibitem{choi2018pact}
Jungwook Choi, Zhuo Wang, Swagath Venkataramani, Pierce I-Jen Chuang,
  Vijayalakshmi Srinivasan, and Kailash Gopalakrishnan.
\newblock Pact: Parameterized clipping activation for quantized neural
  networks.
\newblock {\em arXiv preprint arXiv:1805.06085}, 2018.

\bibitem{bnn}
Matthieu Courbariaux and Yoshua Bengio.
\newblock Binarynet: Training deep neural networks with weights and activations
  constrained to +1 or -1.
\newblock {\em CoRR}, abs/1602.02830, 2016.

\bibitem{deng2012mnist}
Li Deng.
\newblock The mnist database of handwritten digit images for machine learning
  research [best of the web].
\newblock {\em IEEE Signal Processing Magazine}, 29(6):141--142, 2012.

\bibitem{deschamps2012guide}
Jean-Pierre Deschamps, Gustavo~D Sutter, and Enrique Cant{\'o}.
\newblock {\em Guide to FPGA implementation of arithmetic functions}, volume
  149.
\newblock Springer Science \& Business Media, 2012.

\bibitem{deutsch2019generative}
Lior Deutsch, Erik Nijkamp, and Yu Yang.
\newblock A generative model for sampling high-performance and diverse weights
  for neural networks.
\newblock {\em arXiv preprint arXiv:1905.02898}, 2019.

\bibitem{dowdy2011statistics}
Shirley Dowdy, Stanley Wearden, and Daniel Chilko.
\newblock {\em Statistics for research}, volume 512.
\newblock John Wiley \& Sons, 2011.

\bibitem{efron2004least}
Bradley Efron, Trevor Hastie, Iain Johnstone, Robert Tibshirani, et~al.
\newblock Least angle regression.
\newblock {\em The Annals of statistics}, 32(2):407--499, 2004.

\bibitem{googletpu}
Norman P.~Jouppi et. al.
\newblock In-datacenter performance analysis of a tensor processing unit.
\newblock In {\em Proceedings of the 44th Annual International Symposium on
  Computer Architecture, {ISCA} 2017, Toronto, ON, Canada, June 24-28, 2017},
  pages 1--12, 2017.

\bibitem{gallagher1997weight}
Marcus Gallagher and Tom Downs.
\newblock Weight space learning trajectory visualization.
\newblock In {\em Proc. Eighth Australian Conference on Neural Networks,
  Melbourne}, pages 55--59, 1997.

\bibitem{glorot2010understanding}
Xavier Glorot and Yoshua Bengio.
\newblock Understanding the difficulty of training deep feedforward neural
  networks.
\newblock In {\em Proceedings of the thirteenth international conference on
  artificial intelligence and statistics}, pages 249--256, 2010.

\bibitem{go1999analyzing}
Jinwook Go and Chulhee Lee.
\newblock Analyzing weight distribution of neural networks.
\newblock In {\em IJCNN'99. International Joint Conference on Neural Networks.
  Proceedings (Cat. No. 99CH36339)}, volume~2, pages 1154--1157. IEEE, 1999.

\bibitem{han2015learning}
Song Han, Jeff Pool, John Tran, and William Dally.
\newblock Learning both weights and connections for efficient neural network.
\newblock In {\em Advances in neural information processing systems}, pages
  1135--1143, 2015.

\bibitem{he2015delving}
Kaiming He, Xiangyu Zhang, Shaoqing Ren, and Jian Sun.
\newblock Delving deep into rectifiers: Surpassing human-level performance on
  imagenet classification.
\newblock In {\em Proceedings of the IEEE international conference on computer
  vision}, pages 1026--1034, 2015.

\bibitem{he2016deep}
Kaiming He, Xiangyu Zhang, Shaoqing Ren, and Jian Sun.
\newblock Deep residual learning for image recognition.
\newblock In {\em Proceedings of the IEEE conference on computer vision and
  pattern recognition}, pages 770--778, 2016.

\bibitem{ioffe2015batch}
Sergey Ioffe and Christian Szegedy.
\newblock Batch normalization: Accelerating deep network training by reducing
  internal covariate shift.
\newblock {\em arXiv preprint arXiv:1502.03167}, 2015.

\bibitem{islam2020extending}
Muhammad~Aminul Islam, Bryce Murray, Andrew Buck, Derek~T Anderson, Grant~J
  Scott, Mihail Popescu, and James Keller.
\newblock Extending the morphological hit-or-miss transform to deep neural
  networks.
\newblock {\em IEEE Transactions on Neural Networks and Learning Systems},
  2020.

\bibitem{krishnamoorthi2018quantizing}
Raghuraman Krishnamoorthi.
\newblock Quantizing deep convolutional networks for efficient inference: A
  whitepaper.
\newblock {\em arXiv preprint arXiv:1806.08342}, 2018.

\bibitem{krizhevsky2009learning}
Alex Krizhevsky, Geoffrey Hinton, et~al.
\newblock Learning multiple layers of features from tiny images.
\newblock 2009.

\bibitem{krizhevsky2012imagenet}
Alex Krizhevsky, Ilya Sutskever, and Geoffrey~E Hinton.
\newblock Imagenet classification with deep convolutional neural networks.
\newblock In {\em Advances in neural information processing systems}, pages
  1097--1105, 2012.

\bibitem{li2016pruning}
Hao Li, Asim Kadav, Igor Durdanovic, Hanan Samet, and Hans~Peter Graf.
\newblock Pruning filters for efficient convnets.
\newblock {\em arXiv preprint arXiv:1608.08710}, 2016.

\bibitem{merriman1877list}
Mansfield Merriman.
\newblock {\em A List of Writings Relating to the Method of Least Squares: With
  Historical and Critical Notes}, volume~4.
\newblock Academy, 1877.

\bibitem{molchanov2016pruning}
Pavlo Molchanov, Stephen Tyree, Tero Karras, Timo Aila, and Jan Kautz.
\newblock Pruning convolutional neural networks for resource efficient
  inference.
\newblock {\em arXiv preprint arXiv:1611.06440}, 2016.

\bibitem{nakahara2019lightweight}
Hiroki Nakahara, Haruyoshi Yonekawa, Tomoya Fujii, and Shimpei Sato.
\newblock A lightweight yolov2: {A} binarized {CNN} with {A} parallel support
  vector regression for an {FPGA}.
\newblock In {\em Proceedings of the 2018 {ACM/SIGDA} International Symposium
  on Field-Programmable Gate Arrays, {FPGA} 2018, Monterey, CA, USA, February
  25-27, 2018}, pages 31--40, 2018.

\bibitem{rastegari2016xnor}
Mohammad Rastegari, Vicente Ordonez, Joseph Redmon, and Ali Farhadi.
\newblock Xnor-net: Imagenet classification using binary convolutional neural
  networks.
\newblock In {\em European Conference on Computer Vision}, pages 525--542.
  Springer, 2016.

\bibitem{ratzlaff2019hypergan}
Neale Ratzlaff and Li Fuxin.
\newblock Hypergan: A generative model for diverse, performant neural networks.
\newblock {\em arXiv preprint arXiv:1901.11058}, 2019.

\bibitem{redmon2017yolo9000}
Joseph Redmon and Ali Farhadi.
\newblock Yolo9000: better, faster, stronger.
\newblock {\em arXiv preprint}, 2017.

\bibitem{srinivas2017training}
Suraj Srinivas, Akshayvarun Subramanya, and R Venkatesh~Babu.
\newblock Training sparse neural networks.
\newblock In {\em Proceedings of the IEEE Conference on Computer Vision and
  Pattern Recognition Workshops}, pages 138--145, 2017.

\bibitem{tartaglione2018learning}
Enzo Tartaglione, Skjalg Leps{\o}y, Attilio Fiandrotti, and Gianluca Francini.
\newblock Learning sparse neural networks via sensitivity-driven
  regularization.
\newblock In {\em Advances in neural information processing systems}, pages
  3878--3888, 2018.

\bibitem{tibshirani1996regression}
Robert Tibshirani.
\newblock Regression shrinkage and selection via the lasso.
\newblock {\em Journal of the Royal Statistical Society: Series B
  (Methodological)}, 58(1):267--288, 1996.

\bibitem{umuroglu2017finn}
Yaman Umuroglu, Nicholas~J Fraser, Giulio Gambardella, Michaela Blott, Philip
  Leong, Magnus Jahre, and Kees Vissers.
\newblock Finn: A framework for fast, scalable binarized neural network
  inference.
\newblock In {\em Proceedings of the 2017 ACM/SIGDA International Symposium on
  Field-Programmable Gate Arrays}, pages 65--74, 2017.

\end{thebibliography}
}

\end{document}